\documentclass[letterpaper, 10 pt, conference]{ieeeconf}  

\IEEEoverridecommandlockouts                              

\title{\LARGE \bf
Awakening Facial Emotional Expressions in Human-Robot
}

\author{Yongtong Zhu$^{1}$, Lei Li$^{1}$, Iggy Qian$^{2}$, WenBin Zhou$^{3}$, Ye Yuan$^{1}$, Qingdu Li$^{1}$, Na Liu$^{1,*}$, Jianwei Zhang$^{4}$ 
\thanks{$^{1}$Institute of Machine Intelligence, University of Shanghai for Science and Technology, Shanghai 200093, China}%
\thanks{$^{2}$College Of Information Engineering, Zhejiang University of Technology, Hangzhou 310023, China}%
\thanks{$^{3}$Shanghai Droid Robot Co., Ltd. Shanghai 200433, China}%
\thanks{$^{4}$Department of Informatics, University of Hamburg, 20146 Hamburg, Germany}%
\thanks{$^{*}$Corresponding author email: liuna@usst.edu.cn}%
}

\usepackage{graphicx}

\usepackage{amsmath}
\usepackage{algorithm} 
\usepackage{algorithmic} 
\usepackage{float}  
\usepackage{caption} 
\usepackage{amssymb} 
\usepackage{tabularx}
\usepackage{booktabs}
\usepackage{array}

\usepackage[noadjust]{cite}

\begin{document}

\maketitle
\thispagestyle{empty}
\pagestyle{empty}

\begin{abstract}

The facial expression generation capability of humanoid social robots is critical for achieving natural and human-like interactions, playing a vital role in enhancing the fluidity of human-robot interactions and the accuracy of emotional expression. Currently, facial expression generation in humanoid social robots still relies on pre-programmed behavioral patterns, which are manually coded at high human and time costs. To enable humanoid robots to autonomously acquire generalized expressive capabilities, they need to develop the ability to learn human-like expressions through self-training. To address this challenge, we have designed a highly biomimetic robotic face with physical-electronic animated facial units and developed an end-to-end learning framework based on KAN (Kolmogorov-Arnold Network) and attention mechanisms. Unlike previous humanoid social robots, we have also meticulously designed an automated data collection system based on expert strategies of facial motion primitives to construct the dataset. Notably, to the best of our knowledge, this is the first open-source facial dataset for humanoid social robots. Comprehensive evaluations indicate that our approach achieves accurate and diverse facial mimicry across different test subjects.

\end{abstract}


\section{INTRODUCTION}

The ability of humanoid social robots \cite{plutchik2014emotions} to express emotions is crucial, as it not only enhances the naturalness of human-robot interactions but also improves emotional resonance and accuracy during these exchanges. Central to this capability is the skill of facial mimicry \cite{reissland1988neonatal, meltzoff1989imitation, van2009love}, which is essential for humanoid robots to effectively learn and replicate emotional responses. However, the current capabilities of these robots in mimicking 
 expressions are primarily reliant on manually pre-set behavior patterns \cite{oh2006design, hashimoto2006development, hashimoto2008dynamic, ishihara2024automatic}, a method that is both time-consuming and labor-intensive, thereby limiting the robots' flexibility and adaptability.

Current humanoid social robots face substantial challenges in efficiently and adaptively mimicking human facial expressions, primarily due to the absence of a universal learning framework for mastering these skills. Traditional approaches \cite{loza2013application, lin2016expressional, asheber2016humanoid} often rely on predefined behavior patterns to search for the closest matching expressions; however, this method frequently fails to capture the rich diversity of human emotions. Although some frameworks \cite{chen2021smile, hu2024human} based on self-supervised learning exist, they tend to exhibit low learning efficiency and struggle to mitigate the uncanny valley effect, which arises when a robot's appearance and behavior closely resemble human characteristics but remain imperfectly aligned, leading to feelings of discomfort among humans. Consequently, these limitations restrict the practical application of such methods in human-robot interactions, particularly in contexts that demand a high degree of naturalness and accuracy in emotional expression. 

\begin{figure}[t] 
  \centering 
  \includegraphics[width=\linewidth, keepaspectratio]{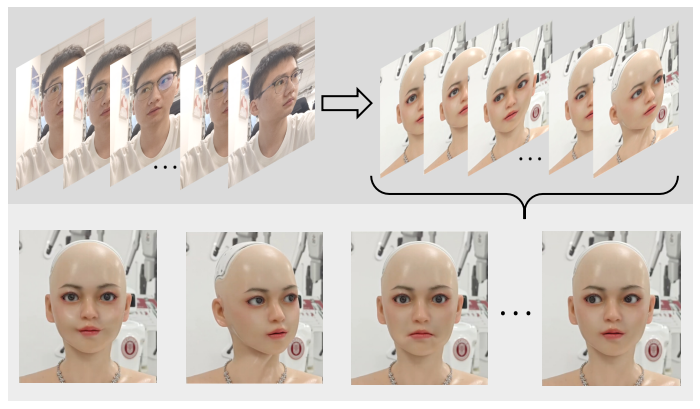}
  \caption{Rena is a general animatronic robotic face for emotional expressions. The robot achieves this by learning the correspondence between its facial feature representations and the control of servos. The entire learning process relies on the robot issuing motor commands based on specific expert strategies. The figure illustrates the robot's capability to replicate a variety of human expressions, with a particular emphasis on the realism of these expressions.}
  \label{fig:introduction_mimic_flow}
  \end{figure}

In this paper, we introduce our robot named Rena, equipped with a 25 DoF control structure and highly biomimetic facial units. Furthermore, we present an innovative end-to-end learning framework based on KAN \cite{liu2024kan} and attention \cite{waswani2017attention} mechanisms that focuses on key visual features to enhance facial imitation learning. Notably, due to the KAN's advantage of low parameter counts, our deployed robotic system achieves an inference speed of up to 50 frames per second. Moreover, we have carefully designed an automated data collection system based on facial motion primitives. Importantly, the constructed dataset aligns naturally with facial motion rules, enabling the model to learn expressions that effectively mitigate the uncanny valley effect.

Our experimental results demonstrate that our method surpasses existing self-supervised learning paradigms in the domain of facial mimicry, as illustrated in Fig.~\ref{fig:introduction_mimic_flow}. We trained our approach on a dataset containing 9,000 samples and achieved a servo control error rate of only 4.4$\%$ on an independent test set. Additionally, we organized a substantial number of participants to evaluate the effectiveness and generalizability of our method. Furthermore, our algorithm exhibits a real-time response time of just 0.02s on the robot, showcasing its exceptional real-time responsiveness, which is critical for interactive control. 

Our contributions can be summarized as follows:
\begin{enumerate}
    \item We present a facial emotion robot featuring highly biomimetic facial units and a high degree-of-freedom facial structure.
    \item We propose an end-to-end learning framework based on KAN and feature attention mechanisms.
    \item We have developed an automated data collection system grounded in expert strategies for facial action primitives.
\end{enumerate}

\section{RELATED WORK}

\subsection{Animatronic Robotics Face}

Designing robots capable of mimicking human expressions serves as a foundational aspect of this work. Previous designs \cite{chen2021smile} typically featured a lower degree of freedom (DoF), which constrained their ability to present diverse facial expressions. Additionally, achieving a variety of expressions while maintaining a natural and lifelike appearance within limited structural space remains a significant challenge \cite{hu2024human}. Recent research on speech-driven animatronic facial expression systems \cite{li2024driving} employs a pneumatically driven system with 16 DoF; however, its limited range restricts expressive capability and lacks head movement flexibility. The Eva robot \cite{chen2021smile}, with 22 DoF, also falls short due to its absence of highly biomimetic facial units, such as artificial eyeballs and skin. In contrast, our robot demonstrates superior expressiveness with 25 DoF combined with advanced biomimetic facial units.

\subsection{Synthetic Video Generation and Animation}

Video synthesis animation is a crucial task in enabling emotional expression for 2D digital humans \cite{peng2023emotalk, zhao2023breathing, liu2024diffdub}, involving the learning of mappings from the source image domain to the target image domain \cite{ma2023dreamtalk, zhang2023dream, liang2024wav2lip}. In contrast, our work focuses on learning the mapping from 2D facial image domains to the target robotic motion space. Recent research on speech-driven animatronic facial expression systems \cite{li2024driving} has implemented a two-stage approach: first driving a 2D digital human face through voice inputs, followed by controlling a robot. However, its mapping, or redirection method, still relies on manually programmed predefined patterns, which limits the generalizability of its expression generation. Our end-to-end learning framework, by contrast, overcomes this limitation, offering a more flexible and generalizable approach to expression generation.

\subsection{Imitation Learning}
Imitation learning, a method that involves learning new tasks by observing and mimicking others' behaviors, has been widely applied in areas such as robotic arm task execution \cite{fu2024mobile, lin2024flowretrieval, hejna2024re, zhao2023learning, zheng2022imitation}. In the field of facial emotion robots, imitation learning enables the robots to learn from human facial expressions, facilitating more natural and human-like emotional interactions. Recent studies utilizing imitation learning frameworks, including Eva 1.0, Eva 2.0, and XIN-REN \cite{chen2021smile, hu2024human, ren2016automatic}, employ a key points displacement tracking approach. However, this method suffers from poor generalization capabilities and is difficult to fine-tune. A significant issue with these approaches is their reliance on random expressions as data sources, which inevitably leads to the uncanny valley effect. In contrast, our robot adopts a different strategy in dataset construction, inherently avoiding the uncanny valley effect. Using a carefully designed facial expression dataset, our method ensures that the learned expressions are natural and highly generalizable.

\section{DESIGN}

\begin{figure}[htbp] 
  \centering 
  \includegraphics[width=\linewidth, keepaspectratio]{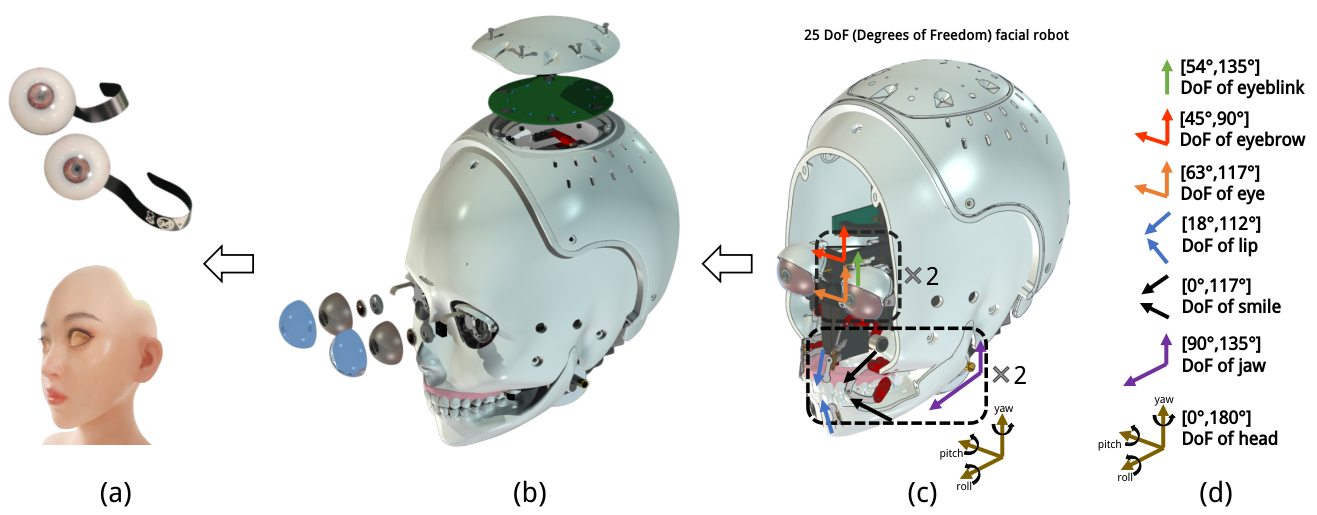}
  \caption{The exploded view of the Rena robot is shown in (b), where (c) is a cross-sectional view of the robot structure, showing the 12 DoF motion trajectory of the mouth motion unit and the 8 DoF motion trajectory of the eyebrow area. (a) shows our bionic silicone skin and bionic eyeball.}
  \label{fig:design}
  \end{figure}

Our robot design, as depicted in Fig.~\ref{fig:design}, incorporates highly realistic biomimetic facial modules and a high-degree-of-freedom structural control module. The structural control module consists of two main components: the head-eye motion unit and the mouth motion unit.

\subsection{Structural Control Module}

The head-eye motion unit is divided into the eyebrow-eye region and the head rotation region. Each eyeball and eyebrow are actuated by two servo motors connected via articulated linkages, which pull the external skin to achieve eyebrow raises, frowns, and horizontal and vertical eye movements.

The mouth motion unit comprises a total of 12 servo motors, which control the upper lip, lower lip, both corners of the mouth, and the jaw opening. Fig.~\ref{fig:design}(c) provides details on the control mechanisms of the mouth servos. In particular, the servo motors for the corners of the mouth and the jaw are coupled for synchronized movement, which poses significant challenges for traditional manual programming of facial expressions.

In the control module, the driving motors exclusively use the FEELECH-STS3032 servo model. This model is a closed-loop servo with a motion range of 180°. Considering the mechanical constraints across the overall structural design, we have additionally imposed specific motion limits on each servo within distinct control regions, as shown in Fig.~\ref{fig:design}(d). This ensures that the actuating mechanisms driven by the various servos avoid interference while generating a wide range of expressions.

\begin{figure*}[htbp] 
  \centering 
  \includegraphics[width=\textwidth, keepaspectratio]{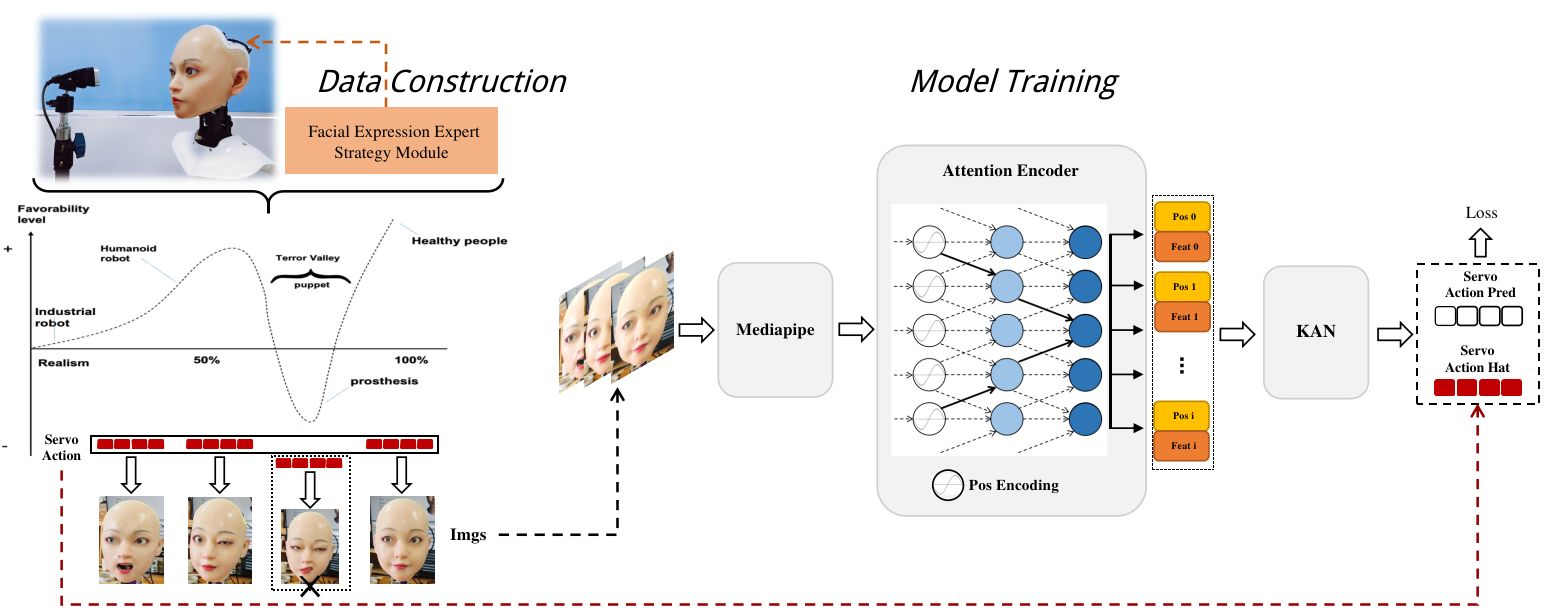}
  \caption{Our method consists of two modules: dataset construction and network training. The dataset construction module automatically generates commands corresponding to random expressions using an expert strategy system, followed by capturing images with a conventional RGB camera. The network training module extracts facial representation blendshape coefficients using MediaPipe. Subsequently, our designed network performs regression to fit the servo commands.}
  \label{fig:method}
  \end{figure*}

\subsection{Biomimetic Module}

Facial emotion robots require not only precise motion structure design but also realistic bionic appearance to promote a more humanoid communication experience. Rena robot achieves this as shown in Fig.~\ref{fig:design}(a). The bionic eyeball we designed achieves unprecedented realism, even replicating the fine details of the blood vessels in the eyeball. This may make us forget that we are communicating with a robot during the interaction. Similarly, the skin is made of bionic silicone through a precise molding process, and its touch is almost the same as that of a real person. Therefore, Rena's appearance, composed of bionic skin and eyeballs, can provide users with an excellent experience during human-robot interaction.

\section{PROPOSED APPROACH}
We propose a learning-based framework for controlling a facial emotion robot to mimic various human facial expressions. First, an automated data collection system is employed to collect data in bulk form. Based on these data, the framework learns the mapping from robot facial expressions to servomotor control commands.

\subsection{Representation of Facial Expression}
We represent facial expressions using the facial blendshape coefficients, a method that research has shown to be scientifically valid and effective \cite{chuang2002performance, mcdonnell2021model, lewis2014practice}. This approach provides a standardized and normalized representation that can be used directly for data learning, as demonstrated in Equation~\ref{eq:bs_defination}. Specifically, we employ MediaPipe, which not only infers facial feature coefficients but also estimates head pose.

\begin{equation}
\label{eq:bs_defination}
S=S_{0}+\sum_{i=1}^{n}\,w_{i}S_{i},
\end{equation}

Where, 
\begin{itemize}
    \item $S$: The final shape of the face.
    \item $S_0$: The base shape (neutral expression).
    \item $S_l$: The blendshape for the $l$-th expression.
    \item $w_l$: The blendshape coefficient for the $l$-th expression, representing the influence of that expression on the final shape.
    \item $n$: The total number of blendshapes, here $n = 52$.
\end{itemize}

\subsection{Dataset Construction}
Self-supervised learning through random facial expressions is an effective approach, as demonstrated by the Eva robot \cite{chen2021smile}. However, in practice, many expressions generated during this process fall within the uncanny valley. To address this issue, we developed an expert strategy system based on our robot's control framework to generate expressions, with the guiding formulation provided in Algorithm~\ref{alg:data_collection}. By following these manually designed rules, the system inherently avoids unnatural expressions at the data distribution level, ensuring more natural and realistic training data. The detail is illustrated in the left part of Fig.~\ref{fig:method}, which provides a comprehensive overview of the data construction phase.

\begin{algorithm}[H] 
\caption{Facial Expression Expert Strategy Module}
\label{alg:data_collection}
\begin{algorithmic}[1]
\STATE \textbf{Define} Servo\_Group = [25 DoF]
\STATE \textbf{Define} Constraints, $\Phi$ = $\{$7 constraint groups$\}$
\STATE \textbf{Define} Emotion\_samples, $S$= $\{\}$

\STATE \textbf{For} iteration $n = 1,2,...$ \textbf{do}
\STATE \quad \textbf{Randomly select} constraint $\phi_i$ from the available 
\STATE \quad constraint groups $\Phi$

\STATE \quad \textbf{Apply constraints:}
\STATE \quad \quad 1. Horizontal eye servos (left-right) synchronized
\STATE \quad \quad 2. Vertical eye servos (up-down) synchronized
\STATE \quad \quad 3. Blink servos synchronized
\STATE \quad \quad 4. Eye opening servos synchronized
\STATE \quad \quad 5. Only one eyebrow movement (frown or raised)
\STATE \quad \quad \quad allowed
\STATE \quad \quad 6. Head movement servos randomly moved
\STATE \quad \quad 7. Mouth servos:
\STATE \quad \quad \quad a. Only one action (smile, sadness, or mouth-
\STATE \quad \quad \quad \quad corners up) allowed
\STATE \quad \quad \quad b. Smile and sadness servos synchronized
\STATE \quad \quad \quad c. Mouth corners up independent

\STATE \quad \textbf{Update} servo positions per selected constraints
\STATE \quad \textbf{Append} current facial image $s_i$ in $S$

\STATE \textbf{End for}
\end{algorithmic}
\end{algorithm}

\subsection{Model Design}

\begin{figure}[H] 
  \centering 
  \includegraphics[width=0.65\linewidth, keepaspectratio]{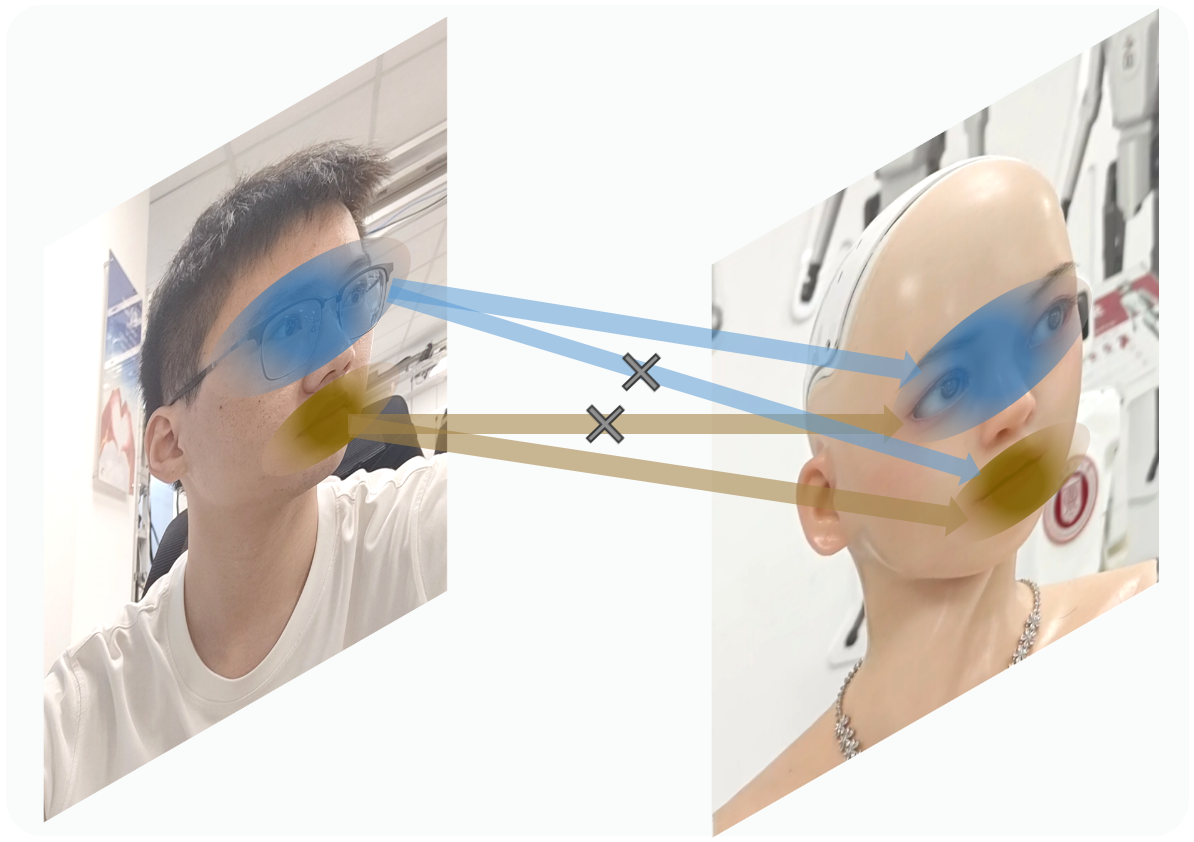}
  \caption{A visual representation of how different facial regions contribute differently to the driving weights of different servos.}
  \label{fig:attention}
  \end{figure}

Previous studies (e.g., Eva \cite{chen2021smile, hu2024human}, XIN-REN \cite{ren2016automatic}) have directly mapped facial features, such as key points, to servo commands. This approach, however, overlooks the fact that different facial regions contribute variably to the driving weights of distinct servos, as illustrated in Fig.~\ref{fig:attention}.

\begin{equation}
PE_{(pos, 2i)} = \sin\left(\frac{pos}{10000^{2i}}\right) \label{eq:pe1}
,
\end{equation}

\begin{equation}
PE_{(pos, 2i+1)} = \cos\left(\frac{pos}{10000^{2i}}\right) 
,
\label{eq:pe2}
\end{equation}

\begin{equation}
Self\_Atten = \text{softmax}\left(\frac{QK^T}{\sqrt{d_k}}\right)\cdot V
,
\label{eq:self_attn}
\end{equation}

Where $i$ and $pos$ represent the blendshape's shape, and $Q$, $K$, and $V$ refer to the input facial features in the self-attention mechanism.

In this study, our model regresses servomotor commands by utilizing blendshape coefficients. Inspired by the variation in driving weights across facial regions, we designed an attention module that enables the model to adaptively select relevant blendshape coefficients. Specifically, we first apply positional encoding to the 52 blendshape coefficients to enhance their distinctiveness. Subsequently, the encoded feature coefficients are processed through a non-linear mapping that incorporates an attention mechanism. The positional encoding and attention formulas are provided in Equations~\ref{eq:pe1}, \ref{eq:pe2}, and \ref{eq:self_attn}. This allows the network to learn the relative importance of various facial features, which is illustrated in the overall workflow diagram shown in the right part of Fig.~\ref{fig:method}.

Considering the importance of response speed in robotic applications, this study employs a KAN instead of an MLP (Multilayer Perceptron) to learn the mapping regression from facial features to servomotor commands. KAN’s learning parameters rely solely on the activation function, and as task complexity increases, the network can select higher-complexity spline activation functions to accommodate the task. Benefiting from KAN’s ability to achieve similar fitting accuracy as MLP with fewer parameters, we integrated it into the overall framework to enhance inference speed. The final fitting formula of the KAN network is presented as $f(x)$, and define $f(x)=KAN(x)$:

\begin{equation}
f(\mathbf{x}) = \sum_{q=1}^{2n+1} \Phi_q \left( \sum_{p=1}^{n} \phi_{q,p}(x_p) \right),
\end{equation}

\text{where } $\phi_{q,p}$ : [0,1] $\rightarrow \mathbb{R}$ \text{ and } $\Phi_q$ : $\mathbb{R} \rightarrow \mathbb{R}$. Specifically, we chose a basic cubic spline curve as the activation function and utilized a 3 layers' KAN network structure as follows:

\begin{equation}
KAN(x) = (\Phi_3 \circ \Phi_2 \circ \Phi_1)(x),
\end{equation}

\begin{equation}
\Phi_i(x) = w_b b(x) + w_s \text{spline}(x).
\end{equation}

We set $b(x) = \text{silu}(x) = \frac{x}{1 + e^{-x}}$, based on the conclusion drawn from KAN, we parameterize the spline function $spline(x)$ as a linear combination of B-splines, such that $ \text{spline}(x) = \sum_i c_i B_i(x)$. This representation allows for a flexible and accurate approximation of complex functions, leveraging the properties of B-splines to ensure smoothness and continuity across the domain of interest.

We also conducted a comparison with the method proposed by Eva \cite{chen2021smile}. Based on their approach, we designed an MLP model that utilizes key points regression to predict servomotor commands. However, directly extracting information from key points is not advisable due to the complexity and abstraction of the data \cite{abiyev2014facial, williams2006performance, park2023said}. The comparative experiments presented later further validate our analysis and demonstrate the effectiveness of our approach.

\subsection{Loss Design}

Considering facial motion during expression generation, we propose an eye movement consistency loss based on our robot's structural control. This loss primarily ensures the consistency of the robot's eye motor movements.

\begin{equation}
L_{con} = |y_{\text{eyeleft}} - y_{\text{eyeright}}| + |y_{\text{browleft}} - y_{\text{browright}}|
,
\end{equation}

\begin{equation}
L_{MSE} = \| \mathbf{y} - \hat{\mathbf{y}} \|_2^2
,
\end{equation}

Where \( y_{\text{eyeleft}} \) represents the command of the left eye, and the others have similar meanings, \( \mathbf{y} \) and \( \hat{\mathbf{y}} \) represent the servomotor commands predicted by the model and the actual servomotor commands.

The main objective function for this task is the MSE (Mean Squared Error) loss, which serves as a performance metric by evaluating the root mean square error between servomotor commands. The total loss function incorporates a hyperparameter $\lambda$, which, after extensive experimental validation, has been optimally set to 0.01. Further details on this experimental determination can be found in the experimental section.

\begin{equation}
L_{total}=L_{MSE}+\lambda L_{con}
,
\end{equation}

\begin{figure*}[t] 
  \centering 
  \includegraphics[width=\textwidth, keepaspectratio]{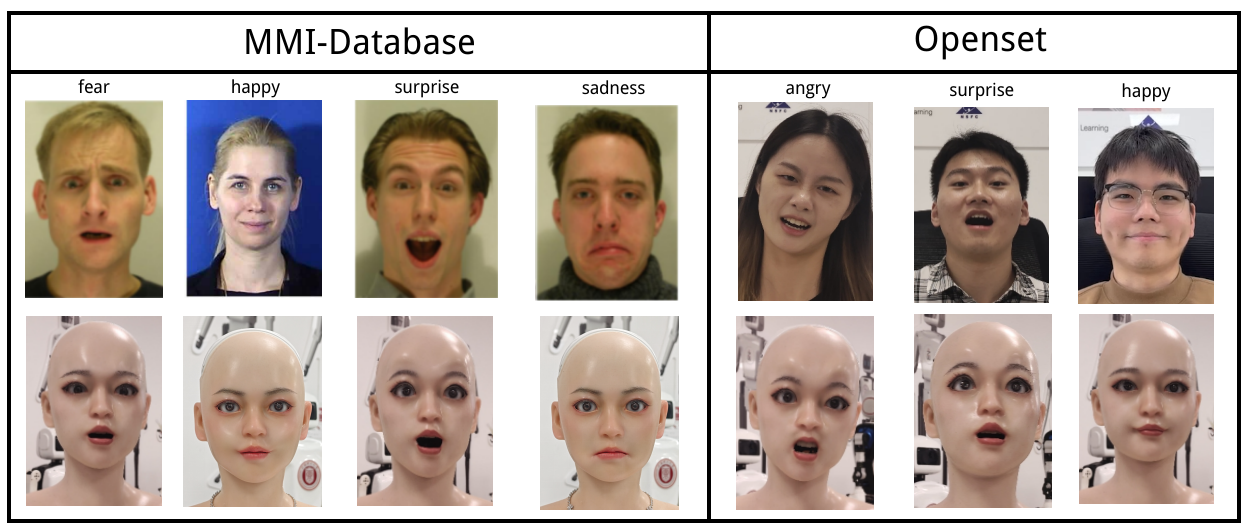}
  \caption{We performed output servomotor commands on our Rena robot on both the MMI dataset and the open dataset to demonstrate that our method supports accurate simulation of various human expressions in multiple human subjects.}
  \label{fig:demo}
  \end{figure*}

\section{EXPERIMENTS}

\subsection{Dataset}

Rena Facial Database: This dataset contains 9,000 sample images of the facial emotion robot, along with the corresponding servomotor commands that drive each expression. Of these, 8,000 images are used for training and 1,000 for testing.

MMI-Database: The MMI database consists of over 2,900 videos and high-resolution static images from 75 subjects, annotated with six basic emotions. We extracted 185 salient facial frames based on these six basic emotions for generalization testing.

Open Set Database: To further evaluate our model's generalization ability, we sampled four volunteers from the lab. Each volunteer participated in comparison experiments by mimicking facial expressions based on the six basic emotions.

The Rena Facial Database is a rigorously designed, comprehensive dataset that allows for detailed analysis of model training and qualitative error analysis, referred to as closed-set experiments. In contrast, the MMI Database and Open Set Database contain unlabeled facial expression samples used for qualitative testing of our model's performance. These samples are analyzed based on specific evaluation metrics and referred to as open-set experiments.

\subsection{Model Training}

\begin{figure}[H] 
  \centering 
  \includegraphics[width=\linewidth, keepaspectratio]{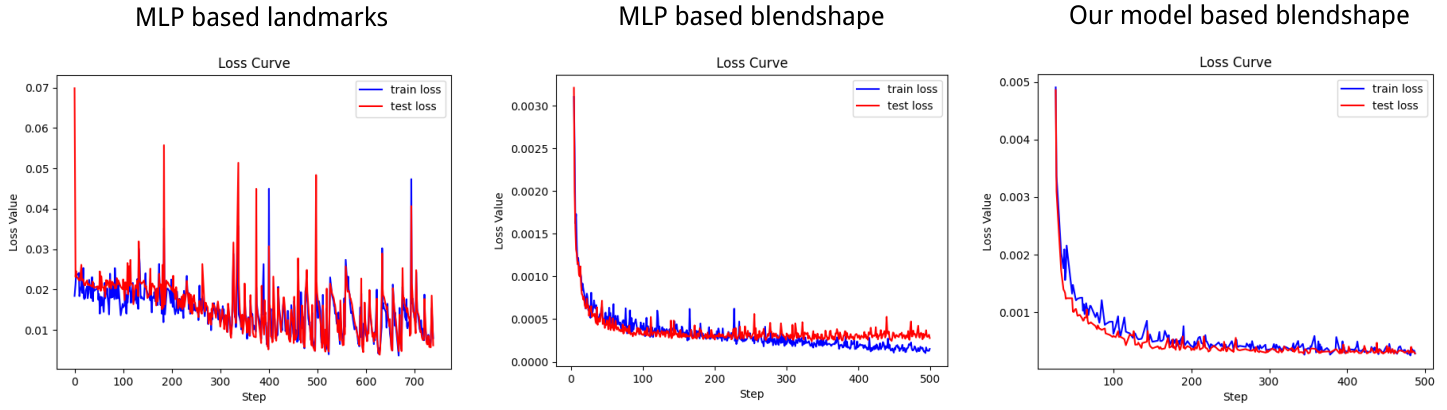}
  \caption{Train loss and test loss during process for the three
mentioned above.}
  \label{fig:loss}
  \end{figure}

On the facial database, we compared three different models: an MLP framework based on facial landmarks, an MLP framework based on facial blendshape coefficients, and our proposed framework. The training loss and testing loss for these three models during the training process are shown in Fig.~\ref{fig:loss}.

\begin{figure}[H] 
  \centering 
  \includegraphics[width=\linewidth, keepaspectratio]{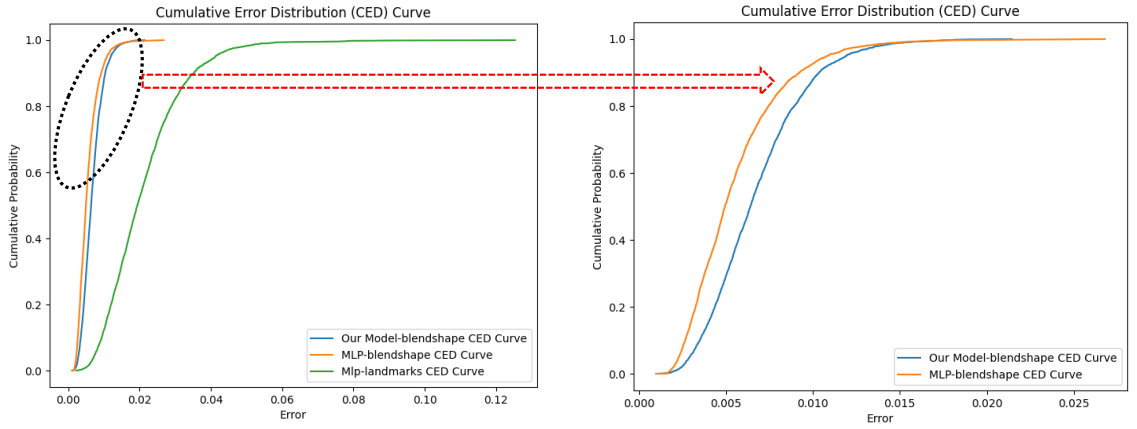}
  \caption{CED curves for the three models mentioned above.}
  \label{fig:ced}
  \end{figure}

The lowest test loss for the three models was 0.03, 0.0025, and 0.0020, respectively. Given that we have normalized the control range of each servo to [0,1], these results indicate average servo control error rates of 17$\%$, 5$\%$, and 4.4$\%$. We also evaluated the overall error distribution of the models. To examine their performance across different error levels, we conducted full inference on the Facial Database using all three models. Subsequently, we assessed their performance using CED (Cumulative Error Distribution) curves, as shown in Fig.~\ref{fig:ced}.

The test results of facial features based on blendshape representation are significantly better than those based on key points representation. Our model, which incorporates the designed KAN and attention mechanism, exhibits the best robustness in handling various facial expressions.

To evaluate the effectiveness of our designed model, we conducted ablation experiments to verify the utility of the attention mechanism. The evaluation metric for these experiments was the generalization error rate on the test set. As shown in Table~\ref{tab:ablation}, the results confirm our hypothesis: incorporating the attention mechanism effectively enhances feature selection capability. Similarly, we designed a series of experiments to verify the efficacy of the proposed consistency loss. The results demonstrate that incorporating consistency loss optimizes model training. Additionally, the trend indicates that as the consistency loss increases from 0.001, the model's optimization performance begins to decrease.

\newcolumntype{C}{>{\centering\arraybackslash}X}
\newcolumntype{L}{>{\raggedright\arraybackslash}X}

\begin{table}[H]
    \centering
    \caption{Ablation Experiments on Hyperparameter $\lambda$ and Attention Mechanism.}
    \label{tab:ablation}
    \begin{tabularx}{\columnwidth}{l C C C}
        \toprule
        \textbf{Backbone} & \textbf{Attention} & \textbf{$\lambda$} & \textbf{Error(\%)} \\
        \midrule
        MLP & \checkmark & - & \textbf{0.0203} \\
        MLP & × & - & 0.0240 \\
        Attention-KAN-bs & \checkmark & 0.0 & 0.0275 \\
        Attention-KAN-bs & \checkmark & 0.1 & 0.0259 \\
        Attention-KAN-bs & \checkmark & 0.01 & \textbf{0.0234} \\
        Attention-KAN-bs & \checkmark & 0.001 & 0.0611 \\
        \bottomrule
    \end{tabularx}
\end{table}

\subsection{Evaluation Metrics}

We deployed our model on the facial emotion robot and visualized the robot's facial expressions alongside the input human facial images. Our evaluation was conducted by comparing 185 salient frames. As shown in the Fig.~\ref{fig:demo}, we present a visual comparison of selected frames between the robot's expressions and the baseline input facial images.

We utilized image distance(ID) and landmark distance(LD) as evaluation metrics, as shown in the following:

\begin{equation}
ID(x, y) = \frac{(2\mu_x \mu_y + C_1)(2\sigma_{xy} + C_2)}{(\mu_x^2 + \mu_y^2 + C_1)(\sigma_x^2 + \sigma_y^2 + C_2)}
,
\end{equation}

\begin{equation}
LD = \frac{1}{n} \sum_{i=1}^{n} \sqrt{(\bar{x}_i - x_i)^2 + (\bar{y}_i - y_i)^2}
,
\end{equation}

In $ID$, $\mu_x$ is the mean value of image $x$. $\sigma_{xy}$ is the covariance of images $x$ and $y$. $C_1$ and $C_2$ are empirical constants, with values of $0.01$ and $0.03$ taken here. In $LD$, $x_i$ and $y_i$ are the coordinates of facial key points.

The former represents pixel-level accuracy, while the latter measures the similarity between facial key points. Our model outperforms the random baseline (RS) and Eva Method by a large margin, as shown in Table~\ref{tab:accuracy}. Note that the image distance is normalized by the total number of pixel values (480, 640, 3), which range from 0 to 1, where the landmark distance is normalized by the 63 landmarks.

\begin{table}[htbp]
\centering
\caption{Accuracy of Different Models}
\label{tab:accuracy}
\begin{tabularx}{0.8\columnwidth}{C C C}
\hline
\textbf{Method} & \textbf{ID} & \textbf{LD} \\
\hline
RS Method & 6.47 & 0.84 \\
Eva Method & 3.47 & 0.46 \\
Our Method & \textbf{2.96} & \textbf{0.074} \\
\hline
\end{tabularx}
\end{table}

Sequential indicators play a crucial role in the subjective experience of humans, as discussed in \cite{yu2014regression}. We further evaluate the performance of an online expression learned from a human performer. This performer executes a variety of facial expressions with neutral-peak-neutral variations. The expression shape vectors of the performer are captured at a rate of 30 frames per second and are used as the robot's target expression vectors.

The values of average space similarity and time similarity reflect the similarity of the imitation trajectory with the performer's facial actions, whereas the value of the average movement smoothness reflects the smoothness of continuous servo motions. The results in Table~\ref{tab:comparison} show the progressiveness of our method compared with three baseline methods using the state-of-the-art humanoid expression generation systems (Jaekel method \cite{jaeckel2008facial}, Trovato method \cite{trovato2013generation}, and Habib method \cite{becker2011evaluating}).

\begin{table}[H]
\centering
\caption{Comparison of Sequential Indicators Versus Four Methods}
\label{tab:comparison}
\begin{tabularx}{\columnwidth}{l C C C}
\hline
\textbf{Sequential indicators} & \textbf{Space similarity} & \textbf{Time similarity} & \textbf{Movement smoothness} \\
\hline
Jaekel Method & 85.4 & 85.2 & 83.3 \\
Trovato Method & 84.4 & 83.9 & 87.6 \\
Habib Method & 87.8 & 86.1 & 84.3 \\
Our Method & \textbf{91.8} & \textbf{88.1} & \textbf{92.7} \\
\hline
\end{tabularx}
\end{table}

The sequential indicators of space-similarity $G_s$, time-similarity $G_t$, and movement smoothness $G_d$, measured by servo hopping during $t_i$ to $t_L$, following the method proposed by Zhu et al. \cite{rawal2022exgennet}, are defined as follows:

\begin{align}
G_s = \frac{1}{n} \sum_{i=1}^{n} \left( \frac{1}{L} \sum_{k=1}^{L} F(d_i^H(t_k) - d_i^R(t_k), b_S) \right)
,
\end{align}

\begin{align}
G_d = 1 - \frac{1}{L} \sum_{k=1}^{L} \frac{1}{m} \sum_{j=1}^{m} G(c_j(t_k))
,
\end{align}

\begin{align}
G_t = &\frac{1}{n} \sum_{i=1}^{n} \left( \frac{1}{L} \sum_{k=1}^{L} F \left( d_i^H(t_k) - d_i^H(t_{k-1}) \right. \right. \notag \\
&\quad \left. - (d_i^R(t_k) - d_i^R(t_{k-1})), b_T \right. \Bigg)
,
\end{align}


Where $L = T = 30$ is the frame rate of camera; $(x_{i0}^H, y_{i0}^H)$ and $(x_{i0}^R, y_{i0}^R)$ represent the $i$th feature point positions of the human and robot at the initial moment, respectively. $d_i^H(t) = \sqrt{(x_i^H(t) - x_{i0}^H)^2 + (y_i^H(t) - y_{i0}^H)^2}$ and $d_i^R(t) = \sqrt{(x_i^R(t) - x_{i0}^R)^2 + (y_i^R(t) - y_{i0}^R)^2}$ are the $i$th feature point displacements of the human and robot at moment $t$, respectively; $F(x, b) = e^{-x^2 / b}$ is a fitting function that converts the deviation parameter $x$ to $(0,1)$ similarity, and $b$ is the parameter used to control the mapping performance. $G(c_j(t_k))$ indicates whether the displacement of the $j$th servo exists unsmoothed hopping at moment $t_k$, and is measured as follows:

\begin{equation}
\begin{aligned}
    & G(c_j(t_k)) = \\
    & \left\{
    \begin{array}{ll}
        1, & \left| c_j(t_k) - c_j(t_{k-1}) \right| - \left| c_j(t_{k-1}) - c_j(t_{k-2}) \right| > T_D, \\
        0, & \text{otherwise}.
    \end{array}
    \right.
\end{aligned}
\end{equation}

\subsection{Implementation Details}
The model was trained using a Titan RTX GPU, with an Intel-F200 camera for data capture. The batch size was set to 256, and the learning rate was 0.00001. The optimizer used for training was ADAM.

\section{CONCLUSIONS}

We propose a facial emotion robot featuring highly biomimetic facial units and a high-degree-of-freedom facial structure. Additionally, we introduce a lightweight end-to-end learning framework specifically designed for facial mimicry. Our experiments demonstrate that our approach can accurately and efficiently recognize facial features and drive the expression robot to replicate similar expressions. Imitation is a crucial step toward endowing robots with more complex skills and serves as a foundational meta-skill for interacting with the external world.

While we show that our robot can successfully imitate various human facial expressions through visual observation, it faces limitations in handling abrupt expression transitions during continuous facial motion. The robot focuses only on the current frame, making it less effective in responding to sudden changes in expressions. Thus, future work should focus on enhancing the model's ability to generate sequential expressions by refining the dataset and the model architecture.

\addtolength{\textheight}{-12cm}  











\bibliography{reference}

\end{document}